\documentclass[11pt]{article}

\PassOptionsToPackage{hyphens}{url}
\usepackage[final]{acl}

\usepackage{times}
\usepackage{latexsym}

\usepackage[T1]{fontenc}

\usepackage[utf8]{inputenc}

\usepackage{microtype}

\usepackage{inconsolata}

\usepackage{graphicx}
\usepackage{booktabs}
\usepackage{multirow}
\usepackage{float}
\usepackage{tikz}
\usetikzlibrary{arrows.meta, positioning, shapes.geometric}
\usepackage{xcolor}
\definecolor{cameragrey}{gray}{0.5}
\newcommand{\camera}[1]{\textcolor{cameragrey}{#1}}
\newenvironment{camerablock}{\par\begingroup\color{cameragrey}}{\par\endgroup}
\renewcommand{\camera}[1]{#1}
\renewenvironment{camerablock}{}{}
\usepackage{amsmath}

\title{HeRE: A Novel Benchmark for Hebrew Relation Extraction via LLM-Guided Denoising of Knowledge Graph Alignments}

\author{
Ron Keinan$^{1}$ \and
Amir David Nissan Cohen$^{2}$ \and
Reut Tsarfaty$^{1}$ \\
$^{1}$Bar-Ilan University \qquad
$^{2}$Origin AI \\
\texttt{\{ronke21, amirdnc, reut.tsarfaty\}@gmail.com}
}

\begin{document}
\maketitle
\begin{abstract}
While LLMs demonstrate strong capabilities in English Relation Extraction (RE) tasks, a significant gap remains in assessing their RE performance in low-resource languages, due to a lack of standardized benchmarks.  
Creating native, language-specific annotated RE benchmarks is expensive and requires human expertise, while relying on translation yields suboptimal, inaccurate datasets that introduce English-centric biases.
To tackle this, we propose a scalable, language-agnostic methodology to create RE data by denoising naturally occurring yet noisy alignments of knowledge bases and corresponding texts. We define a proper alignment via a Natural Language Inference (NLI) task, where a relation triplet is entailed by an input text. 
We empirically assess five denoising strategies, and demonstrate that multilingual models outperform language-specific models in the few-shot LLM setting, while \camera{Hebrew-specific encoders remain competitive under NLI fine-tuning and are strongest in the label-free cross-training setting}. We further find that prompting in the native language achieves performance on par with or slightly exceeding English prompts, confirming that the pipeline operates naturally in the target language. \camera{The best open-weights model, Gemma-4-31B-it (F1\,=\,0.915), matches the best API model, GPT-5.4 (0.921), and reaches F1\,=\,0.844 on a fully held-out, dual-annotated test set.}
Applying the optimized pipeline, we release HeRE: a novel, massive, 1.5-million-instance silver dataset for Hebrew RE, designed to support data-intensive experimentation and model fine-tuning.
HeRE bridges a critical gap in standard Hebrew RE evaluation, while providing the NLP community with a reproducible, language-agnostic pipeline for RE dataset construction in other low-resource settings. 

\end{abstract}

\section{Introduction}
\label{sec:intro}

Relation Extraction (RE) is a fundamental task in Natural Language Processing (NLP), which maps pairs of entity text spans to a predefined set of semantic relations \citep{dixit-al-onaizan-2019-span}, thereby bridging the gap between unstructured text and structured knowledge \citep{zhou-etal-2005-exploring}. 
As a foundational component of Information Extraction, RE is essential for advanced downstream tasks, most notably Knowledge Base construction \citep{zhang-etal-2019-long, niu-etal-2025-tree}, end-to-end information extraction pipelines \citep{keinan2025named}, and Question Answering systems \citep{li-etal-2019-entity, swarup-etal-2025-llm4re}.

In high-resource languages such as English, pre-trained large language models (LLMs) now exhibit state-of-the-art proficiency in both extracting and classifying complex relational triplets \citep{zhao2024comprehensive, xue2024autore, goyal2025named}. 
However, this progress remains heavily confined to high-resource languages. The NLP community critically lacks language-specific standardized benchmarks in low-resource languages \citep{pakray2025natural, Jinensibieke2024HowGA}. Consequently, we lack the empirical resources necessary to evaluate whether the high performance of LLMs achieves true parity in low-resource settings \citep{huguet-cabot-etal-2023-red, ali2025multilingual}.

Overcoming this data bottleneck presents a dual challenge. First, generating natively annotated resources from scratch dictates a heavy reliance on specialized human domain and language expertise. This manual curation is not only financially prohibitive and time-consuming, but it also creates a severe scalability bottleneck, particularly for languages where qualified native annotators are scarce. Second, the common alternative of automatically machine-translating existing English benchmarks often introduces pervasive structural unnaturalness \citep{vanmassenhove-etal-2021-machine, zouhar2026last}. These synthetic pipelines may inject deep-seated linguistic biases, fundamentally failing to capture the authentic cultural context, richness, and semantic nuances inherent to the native target language \citep{zahraei-emami-2025-translate}. To resolve this gap, we argue that the primary scientific challenge is not merely data generation, but rather effectively denoising naturally aligned, real-world information, without relying on exhaustive human annotation or machine translation. 


We suggest a scalable, language-agnostic paradigm. Our approach capitalizes on the naturally occurring alignments between large-scale knowledge bases and their corresponding native text corpus, such as the Wikidata knowledge graph \citep{vrandevcic2014wikidata}, and text sourced from Wikipedia abstracts. While this distant supervision paradigm offers a significant source of authentic relational data without relying on synthetic translation, a vanilla alignment between the text and the knowledge graph exhibits substantial noise. The co-occurrence of two entities in a sentence does not guarantee the expression of their underlying Wikidata relation, leading to a significant prevalence of false positives \citep{huguet-cabot-navigli-2021-rebel-relation}. Therefore, our core objective is to rigorously denoise these noisy natural alignments with a scalable, highly effective pipeline.

To determine the optimal architecture for automated data denoising, we conducted a comprehensive empirical study. We systematically evaluate a broad spectrum of modeling strategies, directly comparing transformer-encoder relation classifiers \citep{liu2022document} and LLMs against encoders and LLMs fine-tuned for Natural Language Inference (NLI), whose entailment judgments are well studied as an annotation task \citep{kalouli-etal-2019-explaining}. By framing text-relation alignment as an entailment task, where the native text acts as the premise and the relational triplet acts as the hypothesis, NLI provides a highly robust, logical mechanism to verify authentic semantic alignment and filter out noisy relations. By evaluating these strategies across both monolingual and multilingual paradigms of models and prompts, we execute a rigorous comparative analysis.

Ultimately, this empirical study demonstrates that multilingual LLMs outperform the competing architectures in the few-shot setting. This advantage comes with important exceptions: \camera{monolingual encoders remain competitive with multilingual encoders under NLI fine-tuning and outperform them when cross-trained on silver relation labels, and task-specific NLI fine-tuning narrows the multilingual advantage for decoder models}.

Applying this methodology, we deliver \textbf{HeRE}, to our knowledge the first comprehensive \textbf{He}brew \textbf{R}elation \textbf{E}xtraction benchmark specifically engineered for Modern Hebrew.\footnote{\camera{Code: \url{https://github.com/Ronke21/HeRE}}}\textsuperscript{,}\footnote{\camera{Data, HeRE-KG and fine-tuned models: \url{https://huggingface.co/datasets/ronke21/HeRE}}} We manually annotated a gold-standard \camera{validation set} of 500 examples, used it to select the denoising method, and applied that method to the remaining noisy alignments to create a silver-standard dataset. HeRE provides a silver-standard dataset of 1.5 million instances, the scale needed to fine-tune Hebrew RE models. \camera{We also release two gold-standard sets of 500 rigorously human-annotated examples each: the validation set on which the pipeline was selected, and a held-out test set, never used for selection, that provides a reliable, culturally grounded benchmark for evaluating LLMs}.

The best open-weights model (Gemma-4-31B-it) reaches F1\,=\,0.915 on the gold validation set, on which all model and prompt choices were made. Its performance is not significantly different from that of the top API model (GPT-5.4, F1\,=\,0.921) under paired bootstrap testing, supporting the feasibility of applying the pipeline to new languages without incurring API costs. Prompting models in Hebrew achieves performance on par with or above English, supporting the pipeline's native-language applicability to native, non-English texts.
To confirm that the denoising quality generalizes beyond the gold benchmark, \camera{we measured how the selected configuration generalizes: on a second, fully held-out test set of 500 dual-annotated examples, on which nothing was tuned, the same model reaches F1\,=\,0.844 overall, and 0.776 on the 203 test examples whose predicates never appeared in the validation set (0.944 on predicates that did).} 

Our contributions are \camera{threefold}. First, we introduce HeRE as a standardized benchmark for Hebrew RE\camera{, together with a quantified comparison to the filter it replaces (+0.131 F1 over CROCODILE's filter after tuning it and +0.509 over its published configuration, for the deployed Gemma-4-31B-it)}. Second, by open-sourcing our validated, language-agnostic dataset generation pipeline, we provide a reproducible blueprint for researchers to systematically generate localized RE benchmarks across the vast spectrum of other low-resource languages. \camera{Third, we demonstrate downstream utility by distilling HeRE into a provenance-grounded Hebrew knowledge graph, showing that, at matched graph size, consensus-denoised edges support better link prediction than randomly selected ones (Sections~\ref{sec:results} and~\ref{sec:here}).}

\section{The Challenge}

Large-scale pre-trained models and, more recently, instruction-tuned LLMs have driven remarkable advancements in NLP throughout the last decade \citep{brown2020language, chang2024survey}. These developments have accelerated progress on high-resource languages, particularly English \citep{qin2026large}, where LLMs now achieve near-human performance on a wide array of information extraction tasks, including RE \citep{wadhwa2023revisiting, zhao2024comprehensive}. However, this progress has not transferred uniformly across languages. Low-resource languages continue to lag significantly behind their high-resource counterparts \citep{joshi-etal-2020-state, hennig2023multitacred,  keinan2023jct, huguet-cabot-etal-2023-red, Jinensibieke2024HowGA, keinan2024text}.

Despite these advances, the low-resource NLP landscape contains a critical gap: there is a lack of native RE benchmarks. We cannot reliably evaluate whether current LLMs, whether monolingual models or multilingual models, have acquired genuine RE competence in low-resource languages. The absence of a large training corpus for RE limits the ability to fine-tune models for Hebrew-specific relation patterns.

The two standard alternatives for creating benchmarks are both problematic at scale. Human annotation from scratch, while gold-standard, is prohibitively expensive at the volume needed for RE, requiring linguistically expert annotators to evaluate thousands of entity-relation-entity triples across diverse document types. Translating existing English benchmarks such as Re-TACRED \citep{stoica-etal-2021-re-tacred} or DocRED \citep{yao-etal-2019-docred} introduces systematic biases: translated texts fail to reflect the cultural and encyclopedic context native to Hebrew Wikipedia, and morphological issues in machine-translated output can corrupt the entity spans that RE models must identify \citep{vanmassenhove-etal-2021-machine}.

We propose a third path that sidesteps both limitations.

\section{The Proposal: Denoising Noisy Text-KG Alignments}

Our starting point is the naturally occurring structural link between two complementary, large-scale resources: the Wikidata knowledge graph \citep{vrandevcic2014wikidata}, which stores billions of structured factual triples of the form $(s, p, o)$, and the Hebrew Wikipedia text corpus, where many of the same facts are expressed in natural language. Wikipedia articles contain hyperlinks to other pages, and each page corresponds to a unique Wikidata entity identifier, providing automatic entity grounding without a separate NER step. Exploiting this correspondence, we align every Hebrew Wikipedia passage in which two entities $s$ and $o$ co-occur to every Wikidata triple $(s, p, o)$ in which they jointly participate, yielding our candidate corpus $\mathcal{C}$. This \textit{distant supervision} paradigm \citep{mintz-etal-2009-distant} provides a scalable, annotation-free source of native-language relational data. Its central drawback is noise: entity co-occurrence is only a weak signal that a passage actually expresses the target relation, so $\mathcal{C}$ contains a substantial fraction of false positives. Denoising $\mathcal{C}$ is therefore the core task we address.

Let $\mathcal{C} = \{(t_i, s_i, p_i, o_i)\}_{i=1}^{N}$ be a corpus of candidate tuples, where $t_i$ is a text passage drawn from a native-language text corpus, $s_i$ and $o_i$ are entity mentions appearing in $t_i$, and $p_i \in \mathcal{P}$ is a relation type from a knowledge graph's predicate vocabulary $\mathcal{P}$. 

The denoising task is to assign a binary label $y_i \in \{0, 1\}$ to each tuple: $y_i = 1$ if and only if $t_i$ explicitly or clearly implicitly asserts that $s_i$ stands in relation $p_i$ to $o_i$, and $y_i = 0$ otherwise. Tuples in $\mathcal{C}$ are produced by distant supervision (any passage where $s_i$ and $o_i$ co-occur is aligned to every knowledge-graph triple $(s_i, p_i, o_i)$ in which they participate), yielding a large but noisy set with an unknown fraction of false positives. Our goal is to learn a denoising function $f\colon \mathcal{C} \rightarrow \{0,1\}^N$ that produces a reliably labeled dataset $\hat{\mathcal{C}} \subseteq \mathcal{C}$.

We build on CROCODILE \citep{huguet-cabot-navigli-2021-rebel-relation} to construct the Hebrew candidate corpus (Appendix~\ref{sec:appendixc}) and replace its single NLI filter with a systematic comparison of five denoising strategies.

\begin{figure*}[!t]
\centering
\resizebox{\linewidth}{!}{%
\begin{tikzpicture}[
  every node/.style={font=\small},
  box/.style={draw, rounded corners=2pt, text centered, inner sep=3pt,
              minimum width=2.1cm, text width=2.0cm, minimum height=1.7cm},
  src/.style={box, fill=blue!10},
  proc/.style={box, fill=gray!14},
  dbox/.style={box, fill=orange!13},
  gbox/.style={box, fill=green!13},
  sbox/.style={box, fill=violet!11},
  arr/.style={-{Latex[length=5pt,width=4pt]}, thick}
]
\node[src]  (kg)    at (0,    1.15) {Knowledge Graph (Wikidata)};
\node[src]  (wiki)  at (0,   -1.15) {Native-Language Wikipedia};
\node[proc] (croc)  at (2.9,  0)    {Text-Triple Alignment};
\node[dbox] (noisy) at (5.8,  0)    {Noisy Triplet Corpus};
\node[gbox] (gold)  at (8.7,  1.15) {\textbf{Gold} Evaluation Sets (2$\times$500)};
\node[dbox] (silr)  at (8.7, -1.15) {Unlabeled Set (2.56M)};
\node[proc] (study) at (11.6, 1.15) {Denoising Study (5 strategies)};
\node[proc] (best)  at (14.5, 1.15) {Optimal Denoiser Selected};
\node[sbox] (silc)  at (17.4, 0)    {\textbf{Annotated Silver Dataset}};
\draw[arr] (kg.east)    -- (croc.west);
\draw[arr] (wiki.east)  -- (croc.west);
\draw[arr] (croc.east)  -- (noisy.west);
\draw[arr] (noisy.east) -- (gold.west);
\draw[arr] (noisy.east) -- (silr.west);
\draw[arr] (gold.east)  -- (study.west);
\draw[arr] (study.east) -- (best.west);
\draw[arr] (best.east)  -- (silc.west);
\draw[arr] (silr.east)  -- (silc.west);
\end{tikzpicture}%
}
\caption{The HeRE construction pipeline.}
\label{fig:pipeline}
\end{figure*}
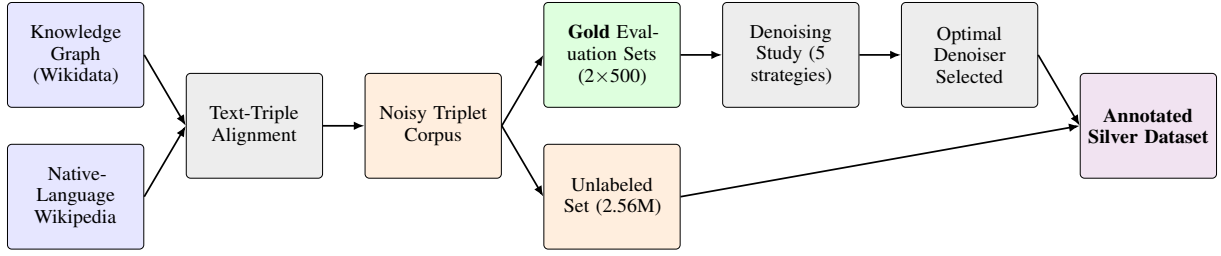

Rather than committing to a single denoising architecture, we adopt a \textbf{multi-strategy approach}: we benchmark five architecturally distinct denoising strategies. The pipeline proceeds in two stages (Figure~\ref{fig:pipeline}): in the \textbf{data preparation stage}, we clean Wikipedia markup and generate three NLI hypothesis variants per triple; in the \textbf{denoising stage}, we apply and compare all five strategies on a gold-standard evaluation set.

\section{Denoising Strategies}
\label{sec:strategies}

We define five strategies for binary KG-text alignment validation. Each takes a tuple $(t, s, p, o)$ and a triple verbalization $h$ (Section~\ref{sec:hyp}) as input and produces a binary prediction $\hat{y} \in \{0,1\}$.

\paragraph{Strategy 1: NLI Encoders.}
A pre-trained encoder is fine-tuned on a Hebrew NLI corpus as a 3-class classifier (entailment\,/\,neutral\,/\,contradiction). At inference time, $t$ is the NLI premise and $h$ is the hypothesis. An entailment-based score ($P(e)$, $P(e){-}P(c)$, or $P(e)/(P(e){+}P(c))$) is thresholded to produce $\hat{y}$. This strategy is fast and GPU-efficient but requires an NLI training corpus transferable to the target language.

\paragraph{Strategy 2: Fine-tuned LLM NLI.}
A decoder LLM is fine-tuned on the same NLI corpus using LoRA \citep{hu-etal-2022-lora}. At inference time it is prompted in NLI format; the binary prediction is derived from first-token log-probabilities of the entailment, neutral, and contradiction label tokens. Fine-tuning adapts the model to the entailment verification framing without requiring gold RE labels.

\paragraph{Strategy 3: Cross-trained Predicate Classifiers.}
Encoder models are trained as multi-class predicate classifiers directly from the raw distant-supervision corpus using $K$-fold cross-training, with no gold labels or NLI data required. For each fold $i$, a model is fine-tuned on the remaining $K{-}1$ folds to predict which predicate $p$ is expressed given $(t, s, o)$ as input. A correct prediction on the held-out fold signals a true positive; an incorrect or uncertain prediction signals a false positive. The final label is majority vote across all $K$ fold-models.

\paragraph{Strategy 4: Open-source LLMs (Few-shot).}
An open-source instruction-tuned LLM is prompted in a binary yes/no format: \emph{does passage $t$ assert relation $p$ between $s$ and $o$?} This is a binary textual entailment question, not a generative extraction task. No fine-tuning is required; prompt language (Hebrew or English) and triple verbalization style are evaluated as independent design choices, with in-context examples provided for calibration.

\paragraph{Strategy 5: Foundation LLMs (Few-shot).}
Commercial frontier models are prompted in the same binary format as Strategy~4, accessed via API. Only hard-label predictions are used; token probabilities are not. These models represent the practical performance ceiling but are cost-prohibitive for large-scale silver annotation.

\section{Experimental Settings}
\label{sec:exp}

We evaluate all five denoising strategies on a common gold-standard benchmark.

\subsection{Data}
\label{sec:data}
Applying CROCODILE to an early-2025 Hebrew Wikipedia dump (\texttt{heb25}) and the matching Wikidata snapshot \camera{yields 3.1 million candidate $(t, s, p, o)$ tuples spanning 1,137 distinct predicates after entity-field normalization. Each tuple pairs a section-level passage from Hebrew Wikipedia with a Wikidata triple where both $s$ and $o$ are mentioned; passage lengths vary considerably (mean 297 words after markup cleaning).}

\subsection{Hypothesis Generation}
\label{sec:hyp}
For each triple we generate three NLI hypothesis variants used across all strategies:
\begin{itemize}
    \item \textbf{Basic}: A surface-form concatenation: \textit{``\{subject\} \{predicate\} \{object\}''}. Straightforward but lacks the grammatical structure of a natural assertion.
    \item \textbf{Template}: A predicate-specific Hebrew sentence from hand-crafted templates covering all predicates in the set. For example, the \textit{place of birth} template produces: \textit{``\{subject\} is the place of birth of \{object\}''}. Templates are grammatically natural and semantically precise.
    \item \textbf{LLM Relation}: A fluent Hebrew sentence generated by Gemma-3-27B-it \citep{team2025gemma3} from the triple, prompted to express the relation in a single natural sentence.
\end{itemize}

\begin{camerablock}
\begin{table}[!t]
\centering
\small
\setlength{\tabcolsep}{3pt}
\begin{tabular}{@{}lrrr@{}}
\toprule
\textbf{Statistic} & \textbf{Gold val.} & \textbf{Gold test} & \textbf{Silver} \\
\midrule
Examples              & 500    & 500    & 1,491,070 \\
Unique documents      & 454    & 494    & $\sim$228K   \\
Triples per document  & 1.1    & 1.0    & $\sim$6.5    \\
Unique predicates     & 97     & 124    & 1,137     \\
Predicates w/ $\geq$20 ex. & 4  & 1      & 593       \\
Positive rate         & 61.4\% & 52.0\% & 72.5\%    \\
Pos.\ rate std (pred.) & 40.6\% & 40.9\% & n/a       \\
Mean text (words)     & 70     & 424    & 297       \\
Mean text (chars)     & 428    & 2,522  & 1,852     \\
\bottomrule
\end{tabular}
\caption{\camera{Dataset statistics.}}
\label{tab:dataset_stats}
\end{table}
\end{camerablock}

\subsection{Models}
\label{sec:models}

For {\bf Strategy~1} we fine-tune six encoders on HebNLI\footnote{\url{https://github.com/NNLP-IL/HebNLI}}: XLM-RoBERTa-large/XL, mE5-large, mmBERT, NeoDictaBERT, and AlephBERT. 

{\bf Strategy~2} fine-tunes nine decoder LLMs on HebNLI: Gemma-2-9B, Gemma3-12B-it, Gemma-3-27B-it, Gemma-4-26B-A4B, Gemma-4-31B, DictaLM-3.0-1.7B-Instruct, DictaLM-3.0-24B, Aya-Expanse-32b, and Qwen3.5-35B-A3B-Base. NLI fine-tuning details and HebNLI performance for all models are reported in Appendix~\ref{sec:appendixd}. 

{\bf Strategy~3} cross-trains XLM-RoBERTa-large, mmBERT, NeoDictaBERT, and mE5-large \camera{with $K \in \{3,5,7,10\}$ (predicates with $<$20 silver examples excluded)}. 

\camera{In {\bf Strategy~4} we evaluate DictaLM-3.0-24B-Base, Gemma-3-27B-it, Gemma-4-26B-A4B-it, Gemma-4-31B-it, and Aya-Expanse-32b at 0/2/5 shots} in Hebrew and English with both hypothesis types. 

In {\bf Strategy~5} we evaluate GPT-5.4, Gemini 3 Flash, and Claude Haiku 4.5\footnote{Accessed through OpenRouter in August 2026 as \path{openai/gpt-5.4}, \path{google/gemini-3-flash-preview}, and \path{anthropic/claude-haiku-4-5}.} at 0 and 5 shots. 
Full setup details are in Appendix~\ref{sec:appendixb}.

\subsection{Evaluation}
\label{sec:metrics}
All strategies are evaluated as binary classifiers on the 500-example gold \camera{validation set (Section~\ref{sec:here}; soft scores are converted to hard predictions as described in Appendix~\ref{sec:appendixb})}. Our primary metric is \textbf{binary F1 on the positive class} (relation present). \camera{F1 is the most informative metric here because the class distribution is imbalanced (61.4\% positive) and both false positives and false negatives carry distinct costs in a denoising pipeline. Because $n=500$ bounds the resolution of any comparison, every reported F1 carries a 95\% confidence interval from a 10{,}000-resample percentile bootstrap, and model pairs are compared with a paired bootstrap on the same examples; the mean CI half-width is $\pm0.022$. Since the positive-class prior varies sharply across predicates (Table~\ref{tab:predicate_rates}, Appendix~\ref{sec:appendix-per-predicate}), we also report macro-F1 averaged over predicates alongside the micro-averaged headline.}

\begin{camerablock}
\subsection{Downstream Evaluation on HeRE-KG}
\label{sec:kge-setup}
To test whether consensus denoising yields measurably better downstream models than the raw alignments it filters, we train TransE~\citep{bordes2013translating} on four variants of the HeRE knowledge graph (Section~\ref{sec:here}) and evaluate filtered link prediction on one shared held-out set of 20,000 triples removed from every training graph, scoring the $\sim$14.2K of them whose entities and relation occur in all four training graphs so that no variant is penalised for unseen vocabulary: \textbf{raw}, all unique distant-supervision triples; \textbf{denoised}, the \texttt{agree3} subset (unanimous positive vote of Gemma-4-31B-it, Gemma-3-27B-it and the HebNLI-tuned DictaLM-3.0-24B; Section~\ref{sec:here}); a \textbf{size-matched random control}, a uniform sample of raw with exactly as many triples as denoised; and \textbf{confidence-weighted}, the full raw graph with training triples sampled in proportion to mean signal confidence. The control separates the effect of selecting better edges from that of simply training on fewer. Held-out triples are sampled from the denoised graph and removed from every training graph; splitting is over unique triples, since a row-level split leaks other mentions of a test triple into training. We report mean and standard deviation over three seeds.
\end{camerablock}

\section{Results}
\label{sec:results}

\begin{table}[!t]
\centering
\small
\setlength{\tabcolsep}{4pt}
\begin{tabular}{llrr}
\toprule
\textbf{Model} & \textbf{Type} & \textbf{F1} & \camera{\textbf{Macro}} \\
\midrule
\multicolumn{4}{l}{\camera{\textit{Baseline: CROCODILE NLI filter}}} \\
\camera{XLM-R-large-XNLI (as published)} & \camera{multi} & \camera{0.406} & \camera{0.299} \\
\camera{XLM-R-large-XNLI (best tuned)} & \camera{multi} & \camera{0.784} & \camera{0.595} \\
\midrule
\multicolumn{4}{l}{\textit{Strategy 1: Fine-tuned NLI Encoders}} \\
AlephBERT              & mono  & 0.618 & \camera{0.560} \\
mE5-large              & multi & \camera{0.692} & \camera{0.614} \\
mmBERT                 & multi & \camera{0.732} & \camera{0.614} \\
XLM-R-large            & multi & \camera{0.756} & \camera{0.680} \\
NeoDictaBERT           & mono  & \camera{0.771} & \camera{0.668} \\
XLM-R-XL               & multi & \camera{\textbf{0.779}} & \camera{0.753} \\
\midrule
\multicolumn{4}{l}{\textit{Strategy 2: Fine-tuned LLM NLI}} \\
\camera{DictaLM-3.0-1.7B-Instruct} & \camera{mono} & \camera{0.429} & \camera{0.259} \\
Gemma-4-26B-A4B        & multi & \camera{0.465} & \camera{0.392} \\
Gemma3-12B-it          & multi & \camera{0.623} & \camera{0.449} \\
Gemma-3-27B-it         & multi & \camera{0.726} & \camera{0.596} \\
DictaLM-3.0-24B-Base   & mono  & \camera{0.803} & \camera{0.676} \\
Aya-Expanse-32b        & multi & \camera{0.829} & \camera{0.702} \\
Gemma-2-9B             & multi & \camera{0.829} & \camera{0.682} \\
\camera{Gemma-4-31B}   & \camera{multi} & \camera{0.849} & \camera{0.760} \\
\camera{DictaLM-3.0-24B} & \camera{mono} & \camera{0.861} & \camera{0.718} \\
\camera{Qwen3.5-35B-A3B-Base} & \camera{multi} & \camera{\textbf{0.873}} & \camera{0.734} \\
\midrule
\multicolumn{4}{l}{\textit{Strategy 3: Cross-trained Predicate Classifiers}} \\
mE5-large              & multi & \camera{0.675} & \camera{0.574} \\
XLM-R-large            & multi & \camera{0.729} & \camera{0.613} \\
mmBERT                 & multi & \camera{0.743} & \camera{0.631} \\
NeoDictaBERT           & mono  & \camera{\textbf{0.751}} & \camera{0.659} \\
\midrule
\multicolumn{4}{l}{\textit{Strategy 4: Open-source LLMs}} \\
DictaLM-3.0-24B-Base   & mono  & \camera{0.835} & \camera{0.717} \\
Aya-Expanse-32b        & multi & \camera{0.847} & \camera{0.732} \\
Gemma-3-27B-it         & multi & \camera{0.877} & \camera{0.726} \\
Gemma-4-26B-A4B-it     & multi & \camera{0.880} & \camera{0.743} \\
Gemma-4-31B-it         & multi & \camera{\textbf{0.915}} & \camera{0.771} \\
\midrule
\multicolumn{4}{l}{\textit{Strategy 5: Foundation LLMs}} \\
Claude Haiku 4.5       & multi & \camera{0.890} & \camera{0.765} \\
Gemini 3 Flash         & multi & \camera{0.917} & \camera{\textbf{0.798}} \\
GPT-5.4                & multi & \camera{\textbf{0.921}} & \camera{0.779} \\
\bottomrule
\end{tabular}
\caption{Best micro-F1 and macro-F1 per model on the gold validation set \camera{(adjudicated labels; macro-F1 over predicates with at least 10 examples)}.}
\label{tab:main_results}
\end{table}

Table~\ref{tab:main_results} reports the best F1 per model on the 500-example
gold \camera{validation set, the set on which all model and configuration choices were made (the held-out test set is reserved for the generalization analysis below)}; complete results appear in
Tables~\ref{tab:full_s1}--\ref{tab:full_s5} (Appendix~\ref{sec:appendixf}).

\begin{camerablock}
\paragraph{HeRE outperforms the CROCODILE baseline it replaces.}
CROCODILE's single XLM-R NLI filter, as published (P(entailment)\,$>$\,0.75 over a surface-concatenated hypothesis), has precision 0.830 but recall 0.269 on Hebrew: it discards two thirds of genuine relations. Tuned, it remains 0.131 below the deployed Gemma-4-31B-it and 0.137 below GPT-5.4, with non-overlapping confidence intervals. Changing only its hypothesis format helps more than retuning only its threshold (+0.32 vs.\ +0.16 F1), corroborating the hypothesis-format finding below; the tuned XNLI model even edges the HebNLI-fine-tuned XLM-R-large of Strategy~1, so for encoders, hypothesis format and threshold matter more than the NLI corpus.
\end{camerablock}

\begin{figure}[!t]
\centering
\includegraphics[width=\linewidth]{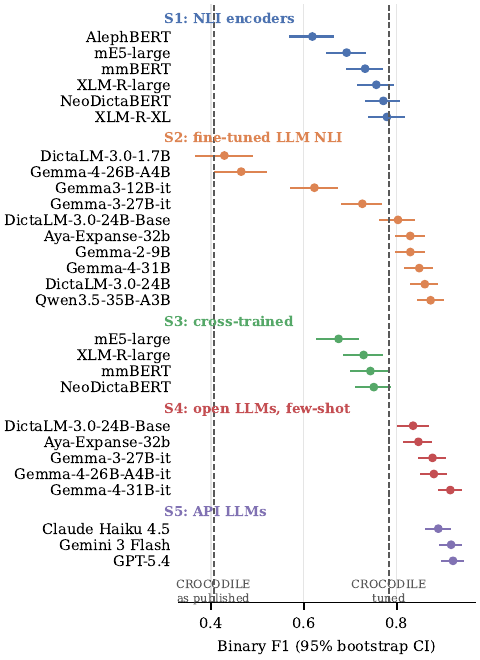}
\caption{\camera{Best micro-F1 per model on the gold validation set with 95\% bootstrap confidence intervals, by strategy. Dashed lines: the CROCODILE filter as published (0.406) and after tuning (0.784).}}
\label{fig:results_ci}
\end{figure}

\paragraph{Open-weights LLMs match the best API models.}
GPT-5.4, Gemini~3~Flash and the open-weights Gemma-4-31B-it \camera{form one statistical cluster (Figure~\ref{fig:results_ci}): a paired bootstrap between the best configuration and the next five (the Hebrew and English Gemini~3~Flash prompts and three Gemma-4-31B-it prompts) finds no distinguishable difference (largest $\Delta=+0.007$, CI $[-0.012,+0.027]$). The best open-weights model is thus not measurably worse than the best commercial API, which is what makes the pipeline reproducible.}

\begin{camerablock}
\paragraph{Macro-F1 reveals much weaker per-predicate performance.}
Across all configurations macro-F1 trails micro-F1 by 0.115 on average, and the ranking changes under macro averaging: Gemini~3~Flash is the most balanced system while Gemma-4-31B-it leads on micro only (Table~\ref{tab:main_results}).

\paragraph{Cross-trained classifiers.}
Cross-training needs neither gold labels nor NLI data but is the costliest strategy ($\sim$480 GPU-hours for the 16-configuration grid, 8--28 per configuration on an A100) and trails Strategy~1's best encoder. NeoDictaBERT leads the multilingual encoders at $K\in\{3,5,7\}$ (by 0.01--0.02, driven by recall) and ties mmBERT at $K{=}10$; F1 varies by at most $\pm0.01$ with $K$, so the number of folds is not a tuning axis (Table~\ref{tab:full_s3}).
\end{camerablock}

\paragraph{Multilingual LLMs win at few-shot prompting; Hebrew encoders win at cross-training.}
The multilingual advantage is task- and scale-dependent. In few-shot prompting (Strategy~4) multilingual models dominate: DictaLM-3.0-24B trails Gemma-4-31B-it by 0.08\camera{, though this comparison is neither size-controlled (24B vs.\ 31B) nor tuning-controlled (a base model against instruction-tuned ones; Appendix~\ref{sec:appendixb})}. For encoders the pattern \camera{reverses or vanishes: XLM-R-XL and NeoDictaBERT are tied in Strategy~1, and NeoDictaBERT leads every multilingual encoder in Strategy~3}. NLI fine-tuning narrows the gap for decoders: \camera{Qwen3.5-35B leads DictaLM-3.0-24B by a margin inside the confidence interval}.

\paragraph{Structured hypotheses outperform plain triple concatenation.}
\label{par:hypothesis-format}
A structured hypothesis beats basic concatenation for every encoder, by up to 0.10 F1 for NeoDictaBERT (Table~\ref{tab:full_s1})\camera{, and the template hypothesis is best for every fine-tuned decoder (Table~\ref{tab:full_s2})}. Few-shot and API models are less sensitive to hypothesis form (Table~\ref{tab:full_s4}).

\paragraph{Prompting in Hebrew works as well as in English.}
The best Hebrew and English configurations are within 0.002 of each other for Gemma-4-31B-it and Gemini~3~Flash, and Hebrew leads by 0.013 for GPT-5.4 (Tables~\ref{tab:full_s4} and~\ref{tab:full_s5}). Frontier LLMs, though trained predominantly on English, are proficient enough in Hebrew for native-language RE classification.

\begin{camerablock}
\paragraph{Denoising improves downstream link prediction.}
\label{sec:kge-results}
At matched training size the denoised graph outperforms the random control by 24\% relative MRR and 29\% relative Hits@10, consistently across seeds (Table~\ref{tab:kge}): the consensus signal selects edges that are genuinely more learnable, not merely fewer. The full raw graph still scores highest in absolute terms, because link prediction rewards coverage; denoising trades coverage for precision. Confidence-weighted sampling keeps raw's coverage but does not improve on uniform sampling (MRR 0.150 vs.\ 0.161), so for link prediction the per-signal scores are best used to choose an operating point, not to reweight training.

\begin{table}[!t]
\centering
\small
\resizebox{\linewidth}{!}{%
\begin{tabular}{lrrr}
\toprule
Training graph & Triples & MRR & Hits@10 \\
\midrule
Raw (all alignments)      & 1{,}479{,}292 & 0.161\,$\pm$\,0.001 & 0.307\,$\pm$\,0.003 \\
Confidence-weighted       & 1{,}479{,}292 & 0.150\,$\pm$\,0.003 & 0.289\,$\pm$\,0.005 \\
\textbf{Denoised (agree3)} & 840{,}991 & \textbf{0.093}\,$\pm$\,0.002 & \textbf{0.189}\,$\pm$\,0.002 \\
Size-matched control      & 840{,}991 & 0.075\,$\pm$\,0.001 & 0.147\,$\pm$\,0.002 \\
\bottomrule
\end{tabular}}
\caption{Link prediction on HeRE-KG. Triples are unique $(s,p,o)$: HeRE-KG's 1,080,605 evidence-linked edges are mentions of 860,990 unique triples, of which 20,000 were drawn for the held-out set and removed from every training graph; the 14,163--14,213 scored per seed are those whose entities and relation appear in every training graph.}
\label{tab:kge}
\end{table}
\end{camerablock}

\begin{camerablock}
\paragraph{Generalization to the held-out test set.}
\label{sec:generalization}
Validation scores are optimistic, since configurations were selected on that set. On the held-out test set (Table~\ref{tab:generalization}) Gemma-4-31B-it loses 0.07 F1 overall and 0.14 on the 203 unseen-predicate examples, while on seen predicates it does not drop at all: the loss is predicate transfer, not overfitting to the annotation scheme. Since the test set is a stress test rather than a sample of the silver distribution (40\% unseen predicates, passages 40\% longer), single-model quality on silver lies between the unseen-predicate and the silver-drawn subset figures. Across every signal deployed on the silver corpus (Table~\ref{tab:test_signals}) the top of the ranking mirrors validation, the seven-signal majority vote is the best rule that labels every row, and the published \texttt{agree3} label is the most accurate on the two thirds it covers (Table~\ref{tab:test_ensembles}), and the gap to the encoder and cross-trained signals widens on these longer passages.

\begin{table}[!t]
\centering
\small
\begin{tabular}{lrr}
\toprule
Evaluation set & $n$ & F1 \\
\midrule
Gold validation (adjudicated)      & 500 & 0.915 \\
\textbf{Gold test (held out)}      & \textbf{500} & \textbf{0.844} \\
\quad seen predicates (depth)      & 97  & 0.944 \\
\quad \textbf{unseen predicates}   & \textbf{203} & \textbf{0.776} \\
\quad silver-validation subset     & 200 & 0.885 \\
\bottomrule
\end{tabular}
\caption{Gemma-4-31B-it: validation set vs.\ held-out test set and its subsets.}
\label{tab:generalization}
\end{table}

\begin{camerablock}
\begin{table}[!t]
\centering
\small
\setlength{\tabcolsep}{3pt}
\begin{tabular}{@{}llrlr@{}}
\toprule
\textbf{Model} & \textbf{Str.} & \textbf{F1} & \textbf{95\% CI} & \textbf{Macro} \\
\midrule
Gemma-4-31B-it & S4 & 0.844 & [0.808, 0.877] & 0.619 \\
Gemma-3-27B-it & S4 & 0.829 & [0.794, 0.862] & 0.543 \\
DictaLM-3.0-24B-Base & S4 & 0.725 & [0.682, 0.766] & 0.426 \\
DictaLM-3.0-24B & S2 & 0.745 & [0.703, 0.785] & 0.454 \\
Gemma-2-9B$^{\ddagger}$ & S2 & 0.332 & [0.271, 0.390] & 0.113 \\
Aya-Expanse-32b$^{\ddagger}$ & S2 & 0.159 & [0.103, 0.217] & 0.132 \\
NeoDictaBERT & S1 & 0.691 & [0.644, 0.734] & 0.419 \\
XLM-R-large & S1 & 0.423 & [0.360, 0.484] & 0.290 \\
NeoDictaBERT ($K{=}3$) & S3 & 0.652 & [0.600, 0.699] & 0.401 \\
mmBERT ($K{=}5$) & S3 & 0.544 & [0.486, 0.600] & 0.372 \\
\bottomrule
\end{tabular}
\caption{Held-out test-set F1 of each signal deployed on the silver corpus. $^{\ddagger}$Deployed at score threshold 0.9. Ensembles: Table~\ref{tab:test_ensembles}.}
\label{tab:test_signals}
\end{table}
\end{camerablock}
\end{camerablock}

\paragraph{Error analysis.}
\camera{Errors fall into three patterns (Appendix~\ref{sec:appendixe}): predicates with no positive gold examples that many configurations still predict, low-prior geographic predicates such as \textit{borders}, and a small set of annotation-ambiguous cases.}

\section{HeRE: The Hebrew Relation Extraction Benchmark}
\label{sec:here}

Drawing on the empirical study above, we release HeRE, the first comprehensive Hebrew RE benchmark, consisting of \camera{three} complementary resources.

\paragraph{Gold datasets.}
We sampled 500 examples from the vanilla-alignment corpus, stratified so that each predicate's share reflects its frequency in the full corpus\camera{, and, to keep annotation tractable, drawn from shorter passages (mean 70 words vs.\ 297 in the corpus; Table~\ref{tab:dataset_stats}), an effect measured in Section~\ref{sec:generalization}}. \camera{Each example was labelled for \textit{relation\_present} (does the passage genuinely express the target Wikidata relation?) independently by two native Hebrew speakers with NLP expertise following the guidelines in Appendix~\ref{sec:appendixa}; inter-annotator agreement is Cohen's $\kappa=0.809$, disagreements were partially adjudicated, and all results are reported against the adjudicated labels. This \emph{validation} set, on which every model and configuration choice was made, covers 97 predicates with 307 positive examples (61.4\%); the standard deviation of per-predicate positive rates is 40.6\% (per-predicate rates in Table~\ref{tab:predicate_rates}, Appendix~\ref{sec:appendix-per-predicate}), underscoring the need for denoising. A second, fully held-out \emph{test} set of 500 examples was dual-annotated in the same way ($\kappa=0.757$) and structured to probe generalization: 203 examples carry predicates absent from the validation set, 97 add depth on seen predicates, and 200 are drawn from the silver distribution. Both sets are released} as a culturally grounded evaluation resource for Hebrew RE systems.

\paragraph{Silver dataset.}
\camera{After removing the 500 validation examples and exact duplicates, 2,564,534 candidate tuples remain as the raw silver corpus. Because API models cannot be applied to 2.5 million examples at acceptable cost, we deploy ten open-weights signals from Strategies~1 to~4 on the full silver corpus (Table~\ref{tab:silver_signals}, Appendix~\ref{sec:appendix-per-predicate})}. To construct the published silver dataset, we require unanimous agreement among \camera{the three strongest open-weights models that were applied to the full corpus (Gemma-4-31B-it and Gemma-3-27B-it from Strategy~4 and DictaLM-3.0-24B from Strategy~2; the Qwen3.5-35B and Gemma-4-26B-A4B-it runs were evaluated on gold only)}, hereafter \texttt{agree3}. This yields \camera{1,080,605 positive examples and 410,465 negative examples (1,491,070 total; 58.1\% of the silver corpus, after withholding the unanimous rows that belong to the held-out test set)}, with a positive rate of 72.5\% within the published set. Per-signal positive rates are given in Table~\ref{tab:silver_signals} (Appendix~\ref{sec:appendix-per-predicate}). Their positive rates range from 54.0\% (Gemma-4-31B-it) to 70.4\% (Gemma-3-27B-it), reflecting calibration differences. \camera{Requiring unanimous agreement across all three models (\texttt{agree3}) retains 58.1\% of silver at a substantially higher confidence threshold. The release is not limited to this binary label: every example ships with the raw score and hard prediction of each of the ten deployed signals and with four consensus rules (Table~\ref{tab:test_ensembles}), so users can choose their own precision/recall operating point or build alternative consensus rules. All 500 test-set rows, which were drawn from the silver corpus, are withheld from every released artifact. The published \texttt{agree3} label reaches precision 0.899, recall 0.934 and F1\,=\,0.916 (Table~\ref{tab:test_ensembles}, Appendix~\ref{sec:appendix-per-predicate}).}

\begin{figure}[!t]
\centering
\includegraphics[width=\linewidth]{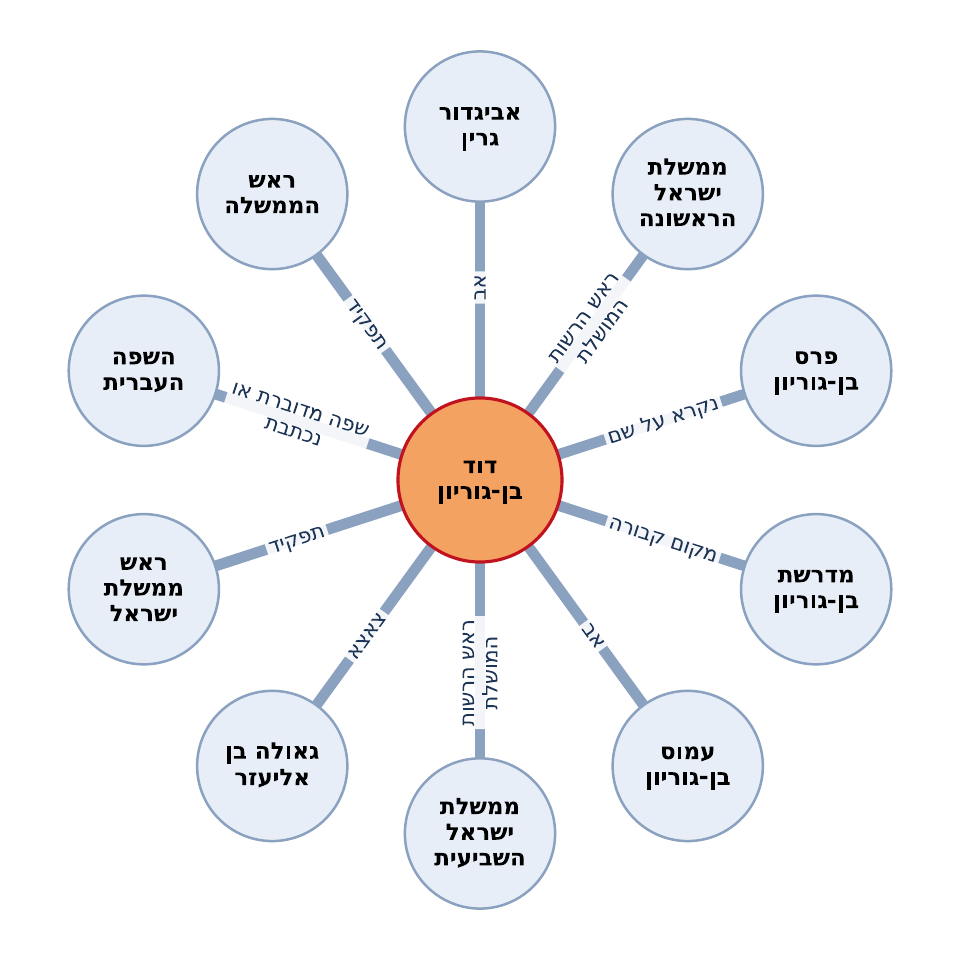}
\caption{\camera{HeRE-KG neighbourhood example: ten highest-confidence edges (width = mean confidence).}}
\label{fig:ego}
\end{figure}

\begin{camerablock}
\paragraph{HeRE-KG.}
We additionally release the positively labelled portion of HeRE as a knowledge graph in which every edge carries the Hebrew Wikipedia passage that expresses it and the per-signal confidence scores behind its label, a textual grounding Wikidata does not provide for Hebrew. HeRE-KG holds 1,080,605 evidence-linked edges over 424,216 entities and 917 relation types (306 with at least 100 edges) from 212,332 documents; 95.6\% of entities lie in one connected component (mean degree 5.1). Figure~\ref{fig:ego} shows one entity's neighbourhood.
\end{camerablock}

\section{Related Work and Future Directions}

\paragraph{Relation Extraction Datasets.}
English RE benchmarking spans sentence-level corpora such as ACE \citep{doddington-etal-2004-automatic} and Re-TACRED \citep{stoica-etal-2021-re-tacred}, document-level DocRED \citep{yao-etal-2019-docred}, and the large-scale REBEL \citep{huguet-cabot-navigli-2021-rebel-relation} with its multilingual successor RED$^{\mathrm{FM}}$ \citep{huguet-cabot-etal-2023-red}, which use the same Wikidata-Wikipedia alignment paradigm we adopt. Despite this progress, native-language RE benchmarks for low-resource languages remain scarce.

\paragraph{Distant Supervision and Denoising.}
Distant supervision for RE was formalized by \citet{mintz-etal-2009-distant} and has underpinned large-scale corpus construction ever since. A central challenge is noise: entity co-occurrence does not guarantee relational expression. Prior denoising approaches include attention-based instance selection \citep{lin-etal-2016-neural} and entailment-based relation verification \citep{sainz-etal-2021-label}. CROCODILE \citep{huguet-cabot-navigli-2021-rebel-relation} applies a single lightweight XLM-RoBERTa NLI filter. Building directly upon their pipeline, we expand this denoising phase into a systematic comparative study of five architectures, establishing multilingual LLMs as the most effective choice in the few-shot setting, while showing that monolingual encoders with NLI fine-tuning remain competitive for low-resource RE benchmark construction.

\paragraph{Multilingual Task Benchmarks.}
Cross-lingual benchmarks exist for dependency parsing \citep{nivre-etal-2016-universal}, morphology \citep{batsuren-etal-2022-unimorph}, question answering \citep{lewis-etal-2020-mlqa, clark-etal-2020-tydi}, and summarization \citep{hasan-etal-2021-xl}. RE remains among the least-resourced multilingual tasks; to our knowledge, HeRE is the first native Hebrew RE benchmark.

\paragraph{Hebrew NLP Resources.}
Hebrew-specific encoders include AlephBERT \citep{seker-etal-2022-alephbert}, DictaBERT \citep{shmidman-etal-2023-dictabert}, and NeoDictaBERT \citep{shmidman2025neodictabert}; the DictaLM family \citep{shmidman-etal-2024-dictalm} provides native Hebrew generative LLMs. Task benchmarks cover NER \citep{bareket-tsarfaty-2021-neural}, morphological parsing \citep{sade-etal-2018-hebrew}, summarization \citep{paz2024hesum}, QA \citep{cohen-etal-2025-hebrewqa}, coreference resolution \citep{greenfeld2026beyond}, and various classification tasks \citep{keinan2025emotional, keinan2026depression}. To our knowledge, Hebrew RE has had no native benchmark prior to this work.

\paragraph{Future Directions.}
The HeRE silver dataset enables end-to-end Hebrew RE model fine-tuning. The language-agnostic pipeline is directly applicable to other languages with sufficient Wikidata-Wikipedia coverage\camera{: its only data prerequisite is a Wikipedia edition linked to Wikidata, which exists for more than 300 languages, most of them low-resource in research attention and LLM pretraining exposure}; extending to Arabic, Persian, and other Semitic languages is a natural next step. \camera{Because HeRE-KG edges are passage-grounded (Section~\ref{sec:here}), they can also be proposed as candidate Wikidata statements for the sparse Hebrew coverage there, each with its supporting passage as the reference Wikidata requires.}

\section{Conclusions}

We have presented HeRE, the first Hebrew Relation Extraction benchmark, together with a language-agnostic pipeline for constructing RE benchmarks via Wikidata-Wikipedia distant supervision and empirically optimized denoising. Across five denoising strategies, multilingual LLMs outperform Hebrew-specific models in the few-shot open-source setting, though monolingual encoders remain competitive\camera{, and in the label-free cross-training setting the Hebrew-specific encoder is the strongest classifier}. The best open-weights model (Gemma-4-31B-it, F1\,=\,\camera{0.915}) \camera{is statistically indistinguishable from} the top commercial API (GPT-5.4, F1\,=\,\camera{0.921}). Hebrew prompting matches English, and \camera{on a fully held-out, dual-annotated test set denoising reaches F1\,=\,0.844 (0.776 on unseen predicates). As a knowledge graph, the corpus yields over a million provenance-grounded edges, and consensus denoising improves link prediction over a size-matched control.} The released datasets\camera{, per-signal scores} and pipeline are a reproducible blueprint for RE benchmarks in any language with a Wikidata-linked Wikipedia.

\begin{camerablock}
\section{Limitations}

\paragraph{Domain scope.}
HeRE is built from Hebrew Wikipedia passages aligned to Wikidata triples, so its text is encyclopedic; models trained or evaluated on it may not transfer to social media, clinical, news or legal Hebrew.

\paragraph{Predicate coverage.}
The gold validation and test sets cover 97 and 124 of the 1{,}137 predicates, few with 20 or more examples: a deliberate consequence of the long-tail setting, which no annotation budget can cover densely. We quantify rather than assume its cost (0.944 on seen vs.\ 0.776 on unseen predicates, \S\ref{sec:generalization}), but F1 on rare predicates remains indicative.

\paragraph{Passage length.}
Gold passages are shorter than silver ones (mean 70 vs.\ 297 words). Stratifying the test set by length, F1 declines from 0.909 (100--250 words) to 0.835 (250--500) and 0.818 (500+; Table~\ref{tab:length_buckets}), a consistent but small trend whose bucket confidence intervals overlap, so length alone does not explain the validation-to-test gap.

\paragraph{Language scope.}
The pipeline is language-agnostic by design but evaluated only on Modern Hebrew; it presumes adequate Wikidata coverage and capable encoders or LLMs, and transfer to a second language, the most direct test of its generality, is left to future work.

\end{camerablock}

\section{Ethical Considerations}
\paragraph{Data Licensing.}
HeRE is derived from Hebrew Wikipedia text and Wikidata, both of which are freely licensed under Creative Commons and open-data licenses respectively. We do not introduce any private or personally sensitive data; all entity and relational information originates from publicly available knowledge bases.

\paragraph{Annotation.}
\camera{Both gold sets were annotated independently by two paid expert annotators compensated at a fair academic rate, with disagreements adjudicated; inter-annotator agreement is reported in Section~\ref{sec:here}. Annotation guidelines were reviewed and refined iteratively to minimize ambiguity (see Appendix~\ref{sec:appendixa}).}

\begin{camerablock}
\section*{Acknowledgments}
This research has been funded by the following funding agencies: a grant from the Israeli Science Foundation (ISF grant 670/23), a KAMIN grant by the Israeli Innovation Authority (IIA), and a VATAT grant by the Israeli Planning and Budgeting Committee (PBC), for which we are grateful.
\end{camerablock}

\bibliography{custom}

\clearpage
\appendix

\section{Annotation Guidelines}
\label{sec:appendixa}

The following guidelines were used by both annotators to assign the binary label \textit{relation\_present} to each example in the gold sets.

\paragraph{Task Definition.}
Given a Hebrew text passage, a subject entity, a Wikidata predicate, and an object entity, the annotator must determine: \textit{Does this passage explicitly or clearly implicitly express the stated relation between the subject and the object?}

\paragraph{Positive (relation present).}
The label is \textbf{1} (relation present) when:
\begin{itemize}
    \item The passage contains a sentence or clause that directly states the relation (e.g., ``X was born in Y'' for a \textit{place of birth} triple).
    \item The relation can be unambiguously inferred from the passage within a single sentence or adjacent sentences, without requiring external knowledge.
\end{itemize}

\paragraph{Negative (relation absent).}
The label is \textbf{0} (relation absent) when:
\begin{itemize}
    \item The passage mentions both entities but does not express the stated relation between them.
    \item The passage expresses a related but distinct relation (e.g., expressing that X \textit{lives in} Y rather than \textit{was born in} Y for a \textit{place of birth} triple).
    \item The relation can only be inferred by combining information across multiple distant passages or requires external knowledge.
    \item The passage is ambiguous as to which of two possible objects the predicate applies.
\end{itemize}

\paragraph{Borderline Cases.}
Annotators were instructed to default to \textbf{0} (relation absent) in genuinely ambiguous cases, and to flag examples where the predicate definition itself was unclear for adjudication.

\section{Experimental Setup}
\label{sec:appendixb}

\paragraph{Hardware.}
Experiments were run on NVIDIA A100-80GB GPUs, with the largest decoders (24B--35B parameters) on B200 nodes and the final Strategy-3 grid cells on V100-32GB nodes. \camera{Encoder NLI models and the cross-trained predicate classifiers were run on a single GPU where memory allowed; the complete cross-training grid (four encoders, $K\in\{3,5,7,10\}$, matched training volume of 2.05M examples per fold) consumed approximately 480 GPU-hours in total (part of the grid ran on slower V100 GPUs), or 8--28 GPU-hours per configuration on an A100 depending on $K$.} Fine-tuned LLM NLI inference (DictaLM-3.0-24B, Aya-Expanse-32b, Gemma-2-9B) used 1--2 A100s via vLLM with LoRA adapter loading. Few-shot open-source LLM experiments (Gemma-3-27B-it, Gemma-4-31B-it) used tensor parallelism across 2 GPUs.

\paragraph{NLI Encoder Fine-tuning.}
Encoder models were fine-tuned on HebNLI for 3-class NLI classification (entailment / neutral / contradiction) with the hyperparameters given in Appendix~\ref{sec:appendixd}, selecting the checkpoint with the best validation macro-F1. The checkpoints used in our experiments are: XLM-RoBERTa-large (checkpoint-8000), XLM-RoBERTa-XL (checkpoint-8000), mE5-large (checkpoint-6000), mmBERT (checkpoint-5800), NeoDictaBERT (checkpoint-5600), AlephBERT (checkpoint-4600).

\paragraph{LLM NLI Fine-tuning.}
LoRA adapters for all nine Strategy-2 decoders were trained on HebNLI with rank-16 LoRA over all linear projection matrices (hyperparameters in Appendix~\ref{sec:appendixd}); the best checkpoint of each model (step 5,280 or 7,920) was used in all NLI inference experiments.

\begin{camerablock}
\paragraph{API models.}
GPT-5.4, Gemini~3~Flash and Claude~Haiku~4.5 were accessed through OpenRouter in August 2026 as \texttt{openai/gpt-5.4}, \texttt{google/gemini-3-flash-preview}, and \texttt{anthropic/claude-haiku-4-5}; request parameters are recorded in the released code.
\end{camerablock}

\begin{camerablock}
\paragraph{Strategy 4 model choice.}
DictaLM-3.0-24B has no instruction-tuned release; we prompt its base model, since its Thinking variant is too costly for corpus-scale prompting. The few-shot comparison between Hebrew-specific and multilingual models is therefore between a base model and instruction-tuned ones.
\end{camerablock}

\begin{camerablock}
\paragraph{Released models.}
Alongside the data, the HeRE collection on the Hugging Face Hub (footnote in Section~\ref{sec:intro}) releases the fine-tuned denoisers needed to regenerate the published signals, together with the per-strategy winners: the HebNLI LoRA adapter of DictaLM-3.0-24B (Strategy~2; one of the three \texttt{agree3} models), the HebNLI-fine-tuned NeoDictaBERT and XLM-R-large encoders (Strategy~1; deployed silver signals), the Strategy-1 winner XLM-R-XL, AlephBERT, and the Strategy-2 adapters of Qwen3.5-35B-A3B-Base and Gemma-4-31B. Each model card records the retained checkpoint, HebNLI development and test macro-F1, and the model's HeRE validation F1. The Strategy-4 signals come from prompting public instruction-tuned models with the prompts in the code release; the Strategy-3 fold models are not released.
\end{camerablock}

\paragraph{Few-shot Prompts.}
The demonstration examples are hand-written and disjoint from both gold sets. Hebrew prompts are literal translations of the English ones below and are included in the released code.

\textit{English triplet prompt (5-shot):}
\begin{quote}
\small
Instruction: Read the passage and determine whether it expresses the given relation. Answer \textbf{yes} if the relation is explicitly stated in the passage, and \textbf{no} otherwise.

[5 demonstration examples]

Passage: \{text\}\\
Relation: \{subject\} -- \{predicate\} -- \{object\}\\
Answer:
\end{quote}
\textit{English template prompt (2-shot):}
\begin{quote}
\small
Instruction: Read the passage and determine whether the following sentence is true according to it. Answer \textbf{yes} or \textbf{no} only.

[2 demonstration examples]

Passage: \{text\}\\
Sentence: \{template\_relation\}\\
Answer:
\end{quote}

\paragraph{Threshold Conversion.}
Our evaluation metrics assume a binary entailed/not-entailed decision per strategy, but most strategies produce soft scores rather than hard labels. We define the conversions as follows. For \textbf{Strategy~1}, we sweep fixed thresholds $\{0.6, 0.7, 0.8, 0.9\}$ across all scoring formulas and hypothesis types and report the best hard-label F1. \camera{(CV threshold optimization tends toward degenerate low-threshold solutions given the 61.4\% positive rate.)} For \textbf{Strategies~2 and~4}, we sweep soft-score thresholds in $\{0.3, 0.4, \ldots, 0.9\}$ and report at the F1-maximizing threshold. For \textbf{Strategy~5}, only hard-label metrics are reported (token probabilities were not used). For \textbf{Strategy~3}, the final prediction is majority vote across $K$ fold-models at threshold 0.5. In all cases, Table~\ref{tab:main_results} reports the best F1 over all hypothesis types and prompt configurations per model.

\section{CROCODILE Data Construction Pipeline}
\label{sec:appendixc}

The CROCODILE pipeline \citep{huguet-cabot-navigli-2021-rebel-relation} constructs a distant-supervision RE corpus for any language with a sufficiently large Wikipedia. The pipeline, on which we base our Hebrew data construction, proceeds as follows.

\paragraph{Step 1: Wikipedia text extraction.}
A Wikipedia dump for the target language is processed using a modified version of WikiExtractor that preserves internal hyperlinks. Each article's text is split into section-level passages (or, optionally, the full article body). Entity surface forms and their Wikipedia page identifiers are identified via the hyperlinks present in the raw Wikipedia markup, providing high-precision entity mention detection without requiring a separate NER system.

\paragraph{Step 2: Wikidata triple retrieval.}
For each Wikipedia entity (subject) identified in Step 1, Wikidata is queried for all relational triples in which it participates as subject. The Wikidata entity identifier is recovered from the Wikipedia page identifier via a Wikipedia-Wikidata mapping. This yields a set of (subject, predicate, object) triples grounded in the Wikidata knowledge graph.

\paragraph{Step 3: Passage-triple alignment.}
For each triple (subject, predicate, object), passages from Step 1 that mention both the subject and the object entity (identified by their Wikipedia hyperlink identifiers) are retrieved and aligned to the triple. The result is a candidate tuple (text, subject, predicate, object) in which the text is the passage where both entities co-occur. Co-occurrence is treated as a weak positive signal that the passage expresses the target Wikidata relation.

\paragraph{Step 4: Initial NLI denoising (CROCODILE default).}
The original CROCODILE pipeline applies an XLM-RoBERTa NLI filter as a lightweight denoising step, discarding candidate tuples for which the passage does not entail a natural-language verbalization of the triple. In our work, we replace and substantially extend this denoising step with the comprehensive empirical denoising study described in Section~\ref{sec:strategies}.

\paragraph{Hebrew-specific notes.}
For the Hebrew heb25 slice, the Wikipedia dump is drawn from Hebrew Wikipedia (early 2025). Entity detection relies on Hebrew Wikipedia's internal hyperlink structure, which is generally reliable for named entities (persons, locations, organizations) but sparser for abstract or relational concepts. Date and numeric value entities are handled by a separate regex-based parser. The resulting heb25 corpus contains 3.1 million candidate tuples spanning 1,257 distinct Wikidata predicate strings before entity-field normalisation (1,137 after; Section~\ref{sec:data}).

\section{HebNLI Fine-tuning}
\label{sec:appendixd}

Strategies~1 and~2 of our denoising pipeline require models pre-trained for Hebrew NLI before being applied to RE. We fine-tune all encoder and decoder models on \textbf{HebNLI}, a Hebrew NLI dataset constructed by translating and adapting MultiNLI and SNLI into Hebrew. HebNLI contains 300,067 training examples, 1,999 validation examples, and 884 test examples across three Hebrew labels, transliterated here as \textit{hisek} (entailment), \textit{stira} (contradiction), and \textit{neitrali} (neutral). The models fine-tuned here are subsequently used as the NLI backbone for our RE denoising experiments; they are not evaluated on RE directly but provide the entailment/contradiction probability estimates that underpin our scoring functions.

\paragraph{Encoder fine-tuning.}
Each encoder is adapted as a 3-class sequence classifier using a linear classification head over the \texttt{[CLS]} representation. Training uses the AdamW optimizer with a learning rate of $2\times10^{-5}$, batch size 64, linear warmup over 10\% of steps, weight decay 0.01, maximum sequence length 250, and a total of 8,000 training steps. Models are evaluated on the validation set every 200 steps; the checkpoint with the best validation macro F1 is retained.

\paragraph{Decoder fine-tuning.}
Decoder models are adapted using LoRA (rank 16, targeting all linear projection matrices, bfloat16 precision) with a learning rate of $5\times10^{-5}$, batch size 16, and up to 8,000 training steps; the checkpoint with the best validation macro-F1 among the periodic evaluations was retained (step 5,280 or 7,920 for every retained model). At inference time, models are prompted with the template \textit{``mishpat 1: \{premise\} mishpat 2: \{hypothesis\}''} (Hebrew for ``sentence 1'' / ``sentence 2'') and the label is extracted from the first generated token.

\paragraph{Results.}
Table~\ref{tab:hebnli_results} reports macro F1 on the HebNLI validation and test sets for all fine-tuned models used in the HeRE denoising experiments. All hardware details are provided in Appendix~\ref{sec:appendixb}.

\begin{table}[t]
\centering
\small
\begin{tabular}{llrr}
\toprule
\textbf{Model} & \textbf{Type} & \textbf{Dev F1} & \textbf{Test F1} \\
\midrule
\multicolumn{4}{l}{\textit{Encoder models}} \\
AlephBERT         & mono  & 0.763 & 0.781 \\
mmBERT            & multi & 0.819 & 0.845 \\
mE5-large         & multi & 0.845 & 0.850 \\
NeoDictaBERT      & mono  & 0.861 & 0.878 \\
XLM-RoBERTa-large & multi & 0.849 & 0.876 \\
XLM-RoBERTa-XL   & multi & 0.853 & 0.882 \\
\midrule
\multicolumn{4}{l}{\textit{Decoder models}} \\
DictaLM-3.0-1.7B-Instruct & mono  & 0.822 & 0.845 \\
Gemma3-12B-it          & multi & 0.851 & 0.886 \\
Qwen3.5-35B-A3B-Base   & multi & 0.869 & 0.904 \\
Gemma-3-27B-it          & multi & 0.857 & 0.894 \\
Gemma-4-26B-A4B            & multi & 0.853 & 0.890 \\
Gemma-4-31B            & multi & 0.870 & 0.905 \\
Gemma-2-9B             & multi & 0.873 & 0.899 \\
Aya-Expanse-32b        & multi & 0.753 & 0.920 \\
DictaLM-3.0-24B-Base & mono  & 0.884 & 0.921 \\
DictaLM-3.0-24B   & mono  & 0.876 & 0.902 \\
\bottomrule
\end{tabular}
\caption{HebNLI macro-F1 of all fine-tuned models.}
\label{tab:hebnli_results}
\end{table}

\section{Error Analysis: Annotated Examples}
\label{sec:appendixe}

Three error categories emerge\camera{, and the per-predicate breakdown (Appendix~\ref{sec:appendix-per-predicate}) exposes the mechanism behind the first two}. First, \textit{official language}, \camera{\textit{language in use}} and \textit{opposite of} \camera{have no positive gold examples under adjudication, because the passages mention both entities while discussing something else, so their F1 is undefined and reported as 0. The best configurations predict at most two false positives on them, but roughly half of all configurations predict the relation at least once}. Second, geographically adjacent predicates, \textit{borders} (\camera{positive rate 0.19}) and \textit{diplomatic relations} (\camera{0.13}), \camera{show the same low-prior signature}: a football tournament or a species listing can mention two neighbouring countries without asserting a border. \camera{The best configurations handle these (F1 0.857 and 1.000) but they remain the main false-positive source for mid-tier models and the origin of the micro-macro gap.} Third, a small residual set is misclassified by every model; several cases involve annotation ambiguity inherent to distant supervision, where a structurally sound triple happens not to be expressed by the aligned passage. 

We present three representative error cases drawn from the gold set, each illustrating a distinct failure mode of the denoising models.

\paragraph{Example 1: False positive from ambiguous co-occurrence (\textit{borders}).}

\noindent\textbf{Subject:} Brazil \quad \textbf{Predicate:} borders \quad \textbf{Object:} Venezuela \quad \textbf{Gold label:} 0 (absent)

\smallskip\noindent\textbf{Passage (translated from Hebrew):}
\begin{quote}\small
The 2000 Copa Libertadores season was the 41st edition of the South American club football competition, organised by CONMEBOL. The first match took place on 15 February 2000 and the competition ended on 21 June 2000\ldots
\end{quote}

\noindent\textbf{Analysis.} Both Brazil and Venezuela appear in the passage as participant nations in a football tournament. The text says nothing about a shared border, yet the geographic plausibility of the relation (Brazil and Venezuela do share a border in Wikidata) causes most models to predict \textit{yes}. This is a prototypical distant-supervision false positive: entity co-occurrence is a weak positive signal that fires even when the passage is semantically unrelated to the target predicate.

\paragraph{Example 2: Annotation ambiguity from incidental evidence (\textit{official language}).}

\noindent\textbf{Subject:} Serbia \quad \textbf{Predicate:} official language \quad \textbf{Object:} Serbian \quad \textbf{Gold label:} 0 (absent)

\smallskip\noindent\textbf{Passage (translated from Hebrew):}
\begin{quote}\small
The Great War Island (in Serbian: \textit{Veliko ratno ostrvo}) is an island in the Danube River, in the Belgrade area, in Serbia. The island was used in many wars, hence its name. The island is located at the confluence of the Sava with the Danube\ldots
\end{quote}

\noindent\textbf{Analysis.} The only hint of the relation is the parenthetical ``in Serbian:'' before the island's local name. The original single-annotator pass labelled this example positive by inferring Serbia's official language from that parenthetical; under the guidelines (Appendix~\ref{sec:appendixa}), a relation that needs external knowledge is negative, and both annotators of the dual-annotation round labelled it 0. All 12 API prompt configurations also answer \textit{no}, while 16 of the 72 open-weights configurations answer \textit{yes}. The case shows how bilingual naming conventions make \textit{official language} look expressed when it is not, which is why this predicate has no positive examples in the adjudicated validation set and why a model that predicts it at all only produces false positives.

\paragraph{Example 3: Universal false positive from predicate granularity (\textit{jurisdiction area}, all models wrong).}

\noindent\textbf{Subject:} Doubs (district) \quad \textbf{Predicate:} jurisdiction area \quad \textbf{Object:} France \quad \textbf{Gold label:} 0 (absent)

\smallskip\noindent\textbf{Passage (translated from Hebrew):}
\begin{quote}\small
Doubs (French: Doubs) is a district in France in the Franche-Comté region, named after the Doubs River. The district was established on 4 March 1790 and is one of the 83 original districts created after the French Revolution\ldots
\end{quote}

\noindent\textbf{Analysis.} The passage explicitly states that Doubs is ``a district in France'', a fact that every model reads as confirming the \textit{jurisdiction area} relation. The annotator, however, labelled this 0: the Wikidata property \textit{applies to jurisdiction} (P1001) is a more specific administrative relation than simple geographic containment, and the passage does not assert it in the formal sense required. This example highlights a broader tension in distant supervision: when Wikidata predicates are semantically finer-grained than the natural-language expressions found in Wikipedia, even accurate models trained on surface co-occurrence will systematically over-predict.

\section{Full Results by Strategy}
\label{sec:appendixf}

Tables~\ref{tab:full_s1}--\ref{tab:full_s5} report complete results for every model and configuration evaluated in this work. The best F1 per model is \textbf{bolded}; these bolded values are the entries reported in Table~\ref{tab:main_results}.

\begin{camerablock}
\begin{table}[H]
\centering
\small
\begin{tabular}{llll}
\toprule
\textbf{Model} & \textbf{Type} & \textbf{Hypothesis} & \textbf{F1} \\
\midrule
\multirow{3}{*}{AlephBERT} & \multirow{3}{*}{mono} & basic & 0.578 \\
 & & template & 0.605 \\
 & & llm & \textbf{0.618} \\
\midrule
\multirow{3}{*}{mE5-large} & \multirow{3}{*}{multi} & basic & 0.642 \\
 & & template & \textbf{0.692} \\
 & & llm & 0.680 \\
\midrule
\multirow{3}{*}{mmBERT} & \multirow{3}{*}{multi} & basic & 0.712 \\
 & & template & \textbf{0.732} \\
 & & llm & 0.723 \\
\midrule
\multirow{3}{*}{XLM-R-large} & \multirow{3}{*}{multi} & basic & 0.685 \\
 & & template & \textbf{0.756} \\
 & & llm & 0.752 \\
\midrule
\multirow{3}{*}{XLM-R-XL} & \multirow{3}{*}{multi} & basic & 0.747 \\
 & & template & \textbf{0.779} \\
 & & llm & 0.740 \\
\midrule
\multirow{3}{*}{NeoDictaBERT} & \multirow{3}{*}{mono} & basic & 0.672 \\
 & & template & \textbf{0.771} \\
 & & llm & 0.736 \\
\bottomrule
\end{tabular}
\caption{Strategy~1: F1 by model and hypothesis, best over scoring formula and fixed threshold.}
\label{tab:full_s1}
\end{table}
\end{camerablock}

\begin{camerablock}
\begin{table}[H]
\centering
\small
\begin{tabular}{llll}
\toprule
\textbf{Model} & \textbf{Type} & \textbf{Hypothesis} & \textbf{F1} \\
\midrule
\multirow{2}{*}{DictaLM-3.0-1.7B-Instruct} & \multirow{2}{*}{mono} & basic & 0.365 \\
 & & template & \textbf{0.429} \\
\midrule
\multirow{2}{*}{Gemma-2-9B} & \multirow{2}{*}{multi} & basic & 0.794 \\
 & & template & \textbf{0.829} \\
\midrule
\multirow{2}{*}{Gemma3-12B-it} & \multirow{2}{*}{multi} & basic & 0.606 \\
 & & template & \textbf{0.623} \\
\midrule
\multirow{2}{*}{Gemma-3-27B-it} & \multirow{2}{*}{multi} & basic & 0.639 \\
 & & template & \textbf{0.726} \\
\midrule
\multirow{2}{*}{Gemma-4-26B-A4B} & \multirow{2}{*}{multi} & basic & 0.440 \\
 & & template & \textbf{0.465} \\
\midrule
\multirow{2}{*}{Gemma-4-31B} & \multirow{2}{*}{multi} & basic & 0.811 \\
 & & template & \textbf{0.849} \\
\midrule
\multirow{2}{*}{DictaLM-3.0-24B-Base} & \multirow{2}{*}{mono} & basic & 0.566 \\
 & & template & \textbf{0.803} \\
\midrule
\multirow{2}{*}{DictaLM-3.0-24B} & \multirow{2}{*}{mono} & basic & 0.789 \\
 & & template & \textbf{0.861} \\
\midrule
\multirow{2}{*}{Aya-Expanse-32b} & \multirow{2}{*}{multi} & basic & 0.793 \\
 & & template & \textbf{0.829} \\
\midrule
\multirow{2}{*}{Qwen3.5-35B-A3B-Base} & \multirow{2}{*}{multi} & basic & 0.825 \\
 & & template & \textbf{0.873} \\
\bottomrule
\end{tabular}
\caption{Strategy~2: best soft-score F1 by model and hypothesis (constrained-prompt reruns for Gemma-4-31B, Qwen3.5-35B, Aya-Expanse-32b and DictaLM-3.0-24B).}
\label{tab:full_s2}
\end{table}
\end{camerablock}

\begin{camerablock}
\begin{table}[H]
\centering
\small
\resizebox{\linewidth}{!}{%
\begin{tabular}{lllrrr}
\toprule
\textbf{Model} & \textbf{Type} & \textbf{K} & \textbf{F1} & \textbf{Macro} & \textbf{Silver pos.\%} \\
\midrule
\multirow{4}{*}{XLM-R-large} & \multirow{4}{*}{multi} & 3 & \textbf{0.729} & 0.613 & 53.5\% \\
 &  & 5 & 0.713 & 0.622 & 56.8\% \\
 &  & 7 & 0.718 & 0.622 & 57.4\% \\
 &  & 10 & 0.714 & 0.621 & 57.9\% \\
\midrule
\multirow{4}{*}{mmBERT} & \multirow{4}{*}{multi} & 3 & 0.732 & 0.614 & 52.1\% \\
 &  & 5 & 0.733 & 0.619 & 54.9\% \\
 &  & 7 & 0.737 & 0.616 & 56.5\% \\
 &  & 10 & \textbf{0.743} & 0.631 & 59.5\% \\
\midrule
\multirow{4}{*}{NeoDictaBERT} & \multirow{4}{*}{mono} & 3 & \textbf{0.751} & 0.659 & 64.3\% \\
 &  & 5 & 0.750 & 0.649 & 65.7\% \\
 &  & 7 & 0.749 & 0.647 & 66.0\% \\
 &  & 10 & 0.742 & 0.645 & 67.5\% \\
\midrule
\multirow{4}{*}{mE5-large} & \multirow{4}{*}{multi} & 3 & \textbf{0.675} & 0.568 & 47.8\% \\
 &  & 5 & 0.674 & 0.576 & 48.4\% \\
 &  & 7 & \textbf{0.675} & 0.574 & 49.6\% \\
 &  & 10 & 0.669 & 0.578 & 50.3\% \\
\bottomrule
\end{tabular}}
\caption{Strategy~3: ensemble F1 and macro-F1 by model and $K$; Silver pos.\% is the ensemble's positive rate on the silver corpus.}
\label{tab:full_s3}
\end{table}
\end{camerablock}

\begin{camerablock}
\begin{table*}[t]
\centering
\small
\setlength{\tabcolsep}{4pt}
\begin{tabular}{lrrrrr}
\toprule
 & & \multicolumn{2}{c}{\textbf{Hebrew F1}} & \multicolumn{2}{c}{\textbf{English F1}} \\
\cmidrule(lr){3-4}\cmidrule(lr){5-6}
\textbf{Model} & \textbf{Shots} & \textbf{triplet} & \textbf{template} & \textbf{triplet} & \textbf{template} \\
\midrule
\multirow{3}{*}{DictaLM-3.0-24B-Base} & 0 & 0.643 & 0.658 & 0.635 & 0.802 \\
  & 2 & 0.822 & 0.818 & 0.723 & 0.811 \\
  & 5 & \textbf{0.835} & 0.822 & 0.595 & 0.821 \\
\midrule
\multirow{3}{*}{Aya-Expanse-32b} & 0 & 0.820 & 0.821 & 0.839 & 0.833 \\
  & 2 & 0.823 & \textbf{0.847} & 0.824 & 0.835 \\
  & 5 & 0.830 & 0.832 & 0.828 & 0.836 \\
\midrule
\multirow{3}{*}{Gemma-3-27B-it} & 0 & 0.860 & 0.875 & 0.844 & 0.860 \\
  & 2 & 0.861 & 0.867 & 0.871 & 0.858 \\
  & 5 & 0.858 & \textbf{0.877} & 0.868 & 0.862 \\
\midrule
\multirow{3}{*}{Gemma-4-26B-A4B-it} & 0 & 0.788 & 0.861 & 0.850 & 0.860 \\
  & 2 & 0.789 & 0.869 & 0.877 & \textbf{0.880} \\
  & 5 & 0.739 & 0.871 & 0.870 & 0.876 \\
\midrule
\multirow{3}{*}{Gemma-4-31B-it} & 0 & 0.868 & 0.915 & 0.851 & 0.905 \\
  & 2 & 0.905 & \textbf{0.915} & 0.913 & 0.910 \\
  & 5 & 0.899 & 0.914 & 0.907 & 0.913 \\
\bottomrule
\end{tabular}
\caption{Strategy~4: F1 by model, hypothesis, shots and prompt language. \textbf{Bold}: best per model.}
\label{tab:full_s4}
\end{table*}
\end{camerablock}

\begin{camerablock}
\begin{table}[H]
\centering
\small
\setlength{\tabcolsep}{4pt}
\begin{tabular}{llrr}
\toprule
\textbf{Model} & \textbf{Lang} & \textbf{Shots} & \textbf{F1} \\
\midrule
\multirow{4}{*}{GPT-5.4} & \multirow{2}{*}{He} & 0 & 0.893 \\
 & & 5 & \textbf{0.921} \\
\cmidrule{2-4}
 & \multirow{2}{*}{En} & 0 & 0.887 \\
 & & 5 & 0.908 \\
\midrule
\multirow{4}{*}{Gemini 3 Flash} & \multirow{2}{*}{He} & 0 & 0.903 \\
 & & 5 & 0.916 \\
\cmidrule{2-4}
 & \multirow{2}{*}{En} & 0 & 0.894 \\
 & & 5 & \textbf{0.917} \\
\midrule
\multirow{4}{*}{Claude Haiku 4.5} & \multirow{2}{*}{He} & 0 & 0.863 \\
 & & 5 & \textbf{0.890} \\
\cmidrule{2-4}
 & \multirow{2}{*}{En} & 0 & 0.850 \\
 & & 5 & 0.889 \\
\bottomrule
\end{tabular}
\caption{Strategy~5: F1 by model, prompt language and shots (triplet hypothesis). \textbf{Bold}: best per model.}
\label{tab:full_s5}
\end{table}
\end{camerablock}

\begin{camerablock}
\section{Per-predicate, Length and Silver-Signal Analyses}
\label{sec:appendix-per-predicate}

\begin{table}[H]
\centering
\small
\setlength{\tabcolsep}{4pt}
\begin{tabular}{lrrrr}
\toprule
 & \multicolumn{2}{c}{\textbf{Validation}} & \multicolumn{2}{c}{\textbf{Test}} \\
\cmidrule(lr){2-3}\cmidrule(lr){4-5}
\textbf{Predicate} & $n$ & \%\,pos. & $n$ & \%\,pos. \\
\midrule
Country                &  79 &  81.0 &  16 &  75.0 \\
Diplomatic relations   &  15 &  13.3 &  29 &   6.9 \\
Instance of            &  29 &  93.1 &   9 &  88.9 \\
Official language      &  19 &   0.0 &  17 &   5.9 \\
Occupation             &  17 &  88.2 &  17 &  94.1 \\
Borders                &  21 &  19.0 &  11 &   0.0 \\
Administrative unit    &  25 &  84.0 &   1 & 100.0 \\
Place of birth         &   7 &  85.7 &  19 &  73.7 \\
Place of origin        &   8 &  12.5 &  16 &   0.0 \\
Citizenship            &  16 &  56.2 &   7 &  85.7 \\
Language in use        &  17 &   0.0 &   5 &   0.0 \\
Capital of             &   8 &  25.0 &  14 &  14.3 \\
Admin.\ subdivision    &  18 &  83.3 &   3 &  33.3 \\
Part of                &  11 &  81.8 &  10 &  70.0 \\
Siblings               &   9 &  66.7 &  12 &  66.7 \\
Subclass of            &  18 &  61.1 &   3 &  33.3 \\
Performer              &  11 &  72.7 &   9 & 100.0 \\
Opposite of            &   9 &   0.0 &  11 &   0.0 \\
Predecessor            &   2 &   0.0 &   7 &  42.9 \\
\bottomrule
\end{tabular}
\caption{\camera{Positive rate of the most frequent predicates in the gold validation and test sets.}}
\label{tab:predicate_rates}
\end{table}

\begin{table}[H]
\centering
\small
\begin{tabular}{lrr}
\toprule
\textbf{Predicate} & $n$ & \textbf{F1} \\
\midrule
Country & 79 & 0.947 \\
Instance of & 29 & 0.962 \\
Administrative unit & 25 & 0.955 \\
Borders & 21 & 0.857 \\
Official language & 19 & n/a$^{*}$ \\
Admin.\ subdivision & 18 & 0.903 \\
Subclass of & 18 & 0.900 \\
Language in use & 17 & n/a$^{*}$ \\
Occupation & 17 & 0.966 \\
Citizenship & 16 & 0.800 \\
Diplomatic relations & 15 & 1.000 \\
Part of & 11 & 0.889 \\
Performer & 11 & 0.947 \\
\bottomrule
\end{tabular}
\caption{Per-predicate F1 of the best configuration (GPT-5.4, Hebrew, triplet, 5-shot) on the validation set, $n\geq10$. $^{*}$No positive gold examples, so F1 is undefined; the model predicts 0 and 2 false positives, respectively.}
\label{tab:per_predicate}
\end{table}

\begin{table}[H]
\centering
\small
\begin{tabular}{lrrl}
\toprule
\textbf{Passage length (words)} & $n$ & \textbf{F1} & \textbf{95\% CI} \\
\midrule
100--250 & 96 & 0.909 & [0.840, 0.966] \\
250--500 & 256 & 0.835 & [0.784, 0.880] \\
500+ & 133 & 0.818 & [0.740, 0.882] \\
\bottomrule
\end{tabular}
\caption{Gemma-4-31B-it test-set F1 by passage length (15 passages under 100 words omitted).}
\label{tab:length_buckets}
\end{table}
\paragraph{Per-signal positive rates and consensus rules.}
The two Strategy-3 signals (NeoDictaBERT $K{=}3$, mmBERT $K{=}5$) predate the grid re-analysis (Table~\ref{tab:full_s3}), which later showed $K$ to be immaterial; they abstain on the 0.4\% of rows whose predicate has fewer than 20 silver examples. Aya-Expanse-32b and Gemma-2-9B were deployed at score threshold 0.9.
\begin{table}[H]
\centering
\small
\setlength{\tabcolsep}{4pt}
\resizebox{\linewidth}{!}{%
\begin{tabular}{llrr}
\toprule
\textbf{Model} & \textbf{Strategy} & \textbf{Pos.} & \textbf{Neg.} \\
\midrule
Gemma-4-31B-it        & S4: Open-source LLM   & 54.0\% & 46.0\% \\
Gemma-3-27B-it        & S4: Open-source LLM   & 70.4\% & 29.6\% \\
DictaLM-3.0-24B-Base  & S4: Open-source LLM   & 66.9\% & 33.1\% \\
DictaLM-3.0-24B       & S2: Fine-tuned NLI    & 67.8\% & 32.2\% \\
Gemma-2-9B            & S2: Fine-tuned NLI    & 23.2\% & 76.8\% \\
Aya-Expanse-32b       & S2: Fine-tuned NLI    & 18.7\% & 81.3\% \\
NeoDictaBERT          & S1: Encoder NLI       & 55.7\% & 44.3\% \\
XLM-R-large           & S1: Encoder NLI       & 35.4\% & 64.6\% \\
NeoDictaBERT ($K{=}3$) & S3: Cross-trained    & 63.1\% & 36.6\% \\
mmBERT ($K{=}5$)      & S3: Cross-trained     & 54.2\% & 45.4\% \\
\midrule
\texttt{agree3} (published) & 3-model unanimous & 72.5\% & 27.5\% \\
\bottomrule
\end{tabular}}
\caption{\camera{Positive rates of the ten deployed signals on the silver corpus; \texttt{agree3} rates are within the published set.}}
\label{tab:silver_signals}
\end{table}

\begin{table}[H]
\centering
\small
\begin{tabular}{lrl}
\toprule
\textbf{Ensemble} & \textbf{F1} & \textbf{95\% CI} \\
\midrule
\texttt{agree3} (published label)$^{\ddagger}$ & 0.916 & [0.886, 0.944] \\
\texttt{vote7} (majority of 7) & 0.801 & [0.763, 0.837] \\
\texttt{agree2} & 0.770 & [0.728, 0.809] \\
\texttt{vote\_all} & 0.680 & [0.628, 0.729] \\
\texttt{agree4} & 0.590 & [0.531, 0.646] \\
\bottomrule
\end{tabular}
\caption{Held-out test-set F1 of consensus rules over the silver-corpus signals: \texttt{agree3} is the unanimous vote of Gemma-4-31B-it, Gemma-3-27B-it and DictaLM-3.0-24B (S2); \texttt{agree2} the unanimous vote of Gemma-3-27B-it and DictaLM-3.0-24B (S2); \texttt{agree4} adds NeoDictaBERT (S1) and NeoDictaBERT $K{=}3$ (S3) to \texttt{agree2}; \texttt{vote7} the majority of the three S4 models, DictaLM-3.0-24B (S2), NeoDictaBERT (S1) and both S3 signals; \texttt{vote\_all} the majority of all ten. Unanimity rules abstain on split votes (abstentions counted as negatives). $^{\ddagger}$; precision 0.899, recall 0.934.}
\label{tab:test_ensembles}
\end{table}

\end{camerablock}

\end{document}